\documentclass{bmvc2k}
\usepackage{colortbl}
\usepackage{xcolor}
\usepackage{multirow}
\usepackage{slashbox}
\usepackage{graphicx}

\title{SAM-RCNN: Scale-Aware Multi-Resolution Multi-Channel Pedestrian Detection}

\addauthor{Tianrui Liu}{t.liu15@imperial.ac.uk}{1}
\addauthor{Mohamed Elmikaty}{melmikat@jaguarlandrover.com}{2}
\addauthor{Tania Stathaki}{t.stathaki@imperial.ac.uk}{1}

\addinstitution{
 Electrical and Electronic Engineering\\
 Imperial College London\\
 London, UK
}
\addinstitution{
 Jaguar Land Rover Research\\
 Coventry, UK
}

\runninghead{Student, Prof, Collaborator}{BMVC Author Guidelines}

\begin{document}

\maketitle

\begin{abstract}
Convolutional neural networks (CNN) have enabled significant improvements in ped-estrian detection owing to the strong representation ability of the CNN features. Recently, aggregating features from multiple layers of a CNN has been considered as an effective approach, however, the same approach regarding  feature representation is used for detecting pedestrians of varying scales. Consequently, it is not guaranteed that the feature representation for pedestrians of a particular scale is optimised. In this paper, we propose a Scale-Aware Multi-resolution (SAM) method for pedestrian detection which can adaptively select multi-resolution convolutional features according to pedestrian sizes. The proposed SAM method extracts the appropriate CNN features that have strong representation ability as well as sufficient feature resolution, given the size of the pedestrian candidate output from a region proposal network. Moreover, we propose an enhanced SAM method, termed as SAM+, which incorporates complementary features channels and achieves further performance improvement. Evaluations on the challenging Caltech and KITTI pedestrian benchmarks demonstrate the superiority of our proposed method.
\end{abstract}

\section{Introduction}
\label{sec:intro}
Pedestrian detection has wide applications in video surveillance, robotics automation and intelligent transportation. A robust pedestrian detector must be able to detect pedestrians of various poses and appearances and at different scales when they are placed in complex scenarios with cluttered backgrounds. A substantial number of methods have been developed in order to improve the detection accuracy \cite{HOG,DPM,Tianrui-DSP2017-Sem,FilterChannal-2015,HowFar-2016,IsfasterRCNNPed,ACF_2014dollar2014,whatcanhelpPed}. In particular, CNN based detectors \cite{SPP_RCNN_2014,fasterRCNN2015,IsfasterRCNNPed} pushed the pedestrian detection performance to a new level. Compared to the traditional feature representation for pedestrian detection, CNN features have strong representation capacity that can largely handle the pose and appearance variations. 

Current CNN-based detection methods use the same CNN feature representation to detect pedestrians of different sizes. 
Nonetheless, a sole feature representation does not always provide the best representation for objects at different sizes. As indicated in \cite{2015scale-awareRCNN}, the visual appearance and the feature representation of large-scaled and small-scaled pedestrians are significantly different. This suggests that there is room for improvement if we could use different feature representation of objects of different sizes.

To solve this problem, we explicitly estimate the scale of the candidate pedestrians and propose to use a \textbf{Scale-Aware Multi-resolution (SAM)} strategy for pedestrian detection. Given the sizes of the candidate pedestrians, we can adaptively select suitable feature representations for pedestrians of different scales, rather than compromising on features that balance for pedestrians of all scales. Our intuition is that the best features (in terms of balancing feature abstraction level and feature resolution) for pedestrians of different sizes may come from different CNN layers. A large-size pedestrian should be represented by features from deep layers, whereas a small-size pedestrian should be represented by features from shallow layers which are of higher resolutions. With the proposed SAM strategy, the detector can choose appropriate CNN features which has the strongest representation ability and meanwhile retains sufficient feature resolution for pedestrians of a specific size.

\begin{figure*}[t]
\begin{centering}
\center\includegraphics[width=1\columnwidth]{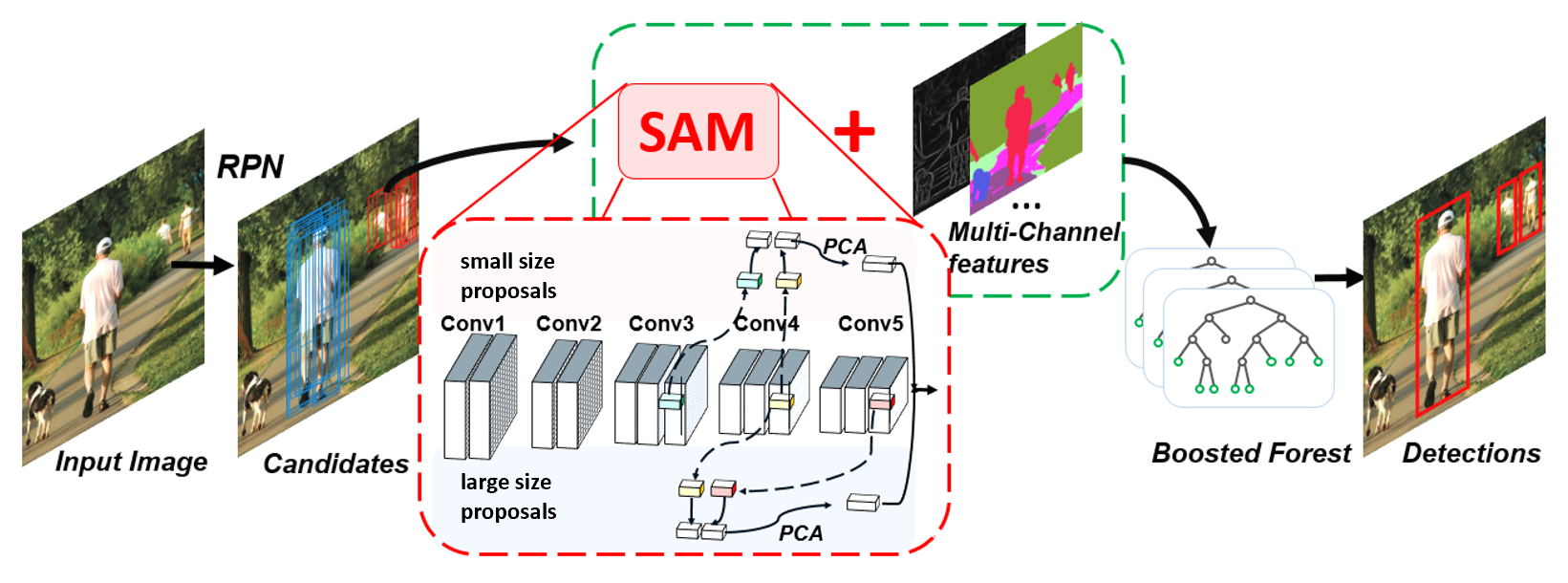}
\par\end{centering}
\caption{Overview of the proposed pedestrian detection framework. Feature extraction is performed differently using the proposed scale-aware multi-resolution (SAM) method according to the pedestrian candidate sizes from RPN (Section 3.2). For small pedestrian candidates, feature maps at shallow layers, such as $Conv3$ and $Conv4$, are used, while for large pedestrian candidates, feature maps at deeper layers are utilized because the feature resolution is sufficient large. SAM+ uses complementary feature channels with the CNN features (Section 3.3) }
\vspace*{-0.5em}
\label{fig:long-framework}
\end{figure*}

Furthermore, we propose an enhanced version of the SAM detector, denoted as \textbf{SAM+}, by incorporating complementary feature channels with the CNN features for a better pedestrian detection. A novel \textbf{RoI histogram pooling} method is proposed to extract feature vectors from the additional feature maps for candidate regions of arbitrary size, leading to better performance than using RoI max pooling \cite{fasterRCNN2015}. With the complementary features, our SAM+ detector can effectively eliminate hard negatives such as tree leaves and traffic lights. 
As pedestrians of different size may use different combinations of CNN layers and thus have features of different dimensions, we apply principal component analysis (PCA) to transform the feature vector into a fixed length vector such that it can be easily fed into the boosted forest (BF) classifier \cite{RealBoost1999}. 
The flexibility of the BF classifier imposes no need for feature amplitudes normalization when combining multi-resolution CNN features and facilitates the integration of additional feature channels for the SAM+ detector. 

The contributions of this work can be summarized as follows: 
1. We proposed a scale-aware multi-resolution pedestrian detection framework which exploits multi-resolution features from CNN and uses different combination of feature layers according to the size of the candidate pedestrians. 
2. We propose an enhanced version of the SAM, termed as SAM+, which uses additional semantic and edge feature channels to obtain valuable complementary information for pedestrian detection. Experimental results show improvements on detection rates using additional cues. 
3. The proposed SAM pedestrian detector achieves the state-of-the-art on the Caltech and KITTI pedestrian benchmarks. Along with the enforcement of the additional features channels, the proposed SAM+ detector outperforms the state-of-the-art.
\vspace{-1.44em}

\section{Related Work}

Traditional pedestrian detectors, such as ACF \cite{ACF_2014dollar2014} and Checkerboards \cite{FilterChannal-2015}, are based on hand-engineered features which are usually descriptors of the gradients, edge and colors computed over a sliding window. These descriptors are used in conjunction with a classifier, such as boost forest, to perform pedestrian detection via classification. These methods were the dominant approaches for pedestrian detection before the emerging of the CNN based pedestrian detection methods.

Compared to the hand-engineered features, features generated by deep convolutional networks have demonstrated to have stronger feature representation capability in many detection tasks. Region-based convolutional neural networks (R-CNN) \cite{RegionCNN-2014CVPR} method makes use of an "attention" mechanism which first proposes a small number of high potential candidate regions where classification is performed afterwards. R-CNN \cite{RegionCNN-2014CVPR} and Fast R-CNN \cite{fastrcnn15} apply selective search \cite{SelectiveSearch2013} for region proposal, and Faster R-CNN \cite{fasterRCNN2015} replaces selective search with a built-in region proposal network (RPN) network that can effectively generate proposals. These general object detection methods when directly applied for pedestrian detection do not lead to satisfying results \cite{IsfasterRCNNPed,CityPersons} because Faster-RCNNs do not perform well on small size objects which dominate most pedestrian datasets \cite{PedBenchmark-2009CVPR-Dollar,KITTI-data}. In \cite{IsfasterRCNNPed}, the Faster R-CNN is tailored to accommodate for pedestrian detection and achieves better results.

In an attempt to take advantage of the CNN features from multiple layers, the Inside-Outside Net \cite{bell2016inside} concatenates multiple layers of CNN and unifies the feature dimensions using a $1\times1$ convolution. As the convolutional features at each layer have very different amplitudes, they need to rely on $L_2$ normalization to normalize the features before the concatenation. In single shot multi-box detection (SSD) \cite{MultiBox2016ssd}, several additional layers are built after $Conv5\_3$ of the VGG16 net \cite{VGG16}. SSD combines the CNN features starting from $Conv5\_3$ with the additional layers. However, SSD does not use the higher-resolution maps before $Conv5\_3$ that are crucial for detecting small objects. Feature pyramid network \cite{FPN2016} combines low-resolution with high-resolution feature maps via a top-down pathway. The lower resolution feature maps are up-sampled by a factor of $2$ and merged with the upper layer feature map by element-wise addition. The result is a feature pyramid that has high resolution and strong representation ability at all levels. Although these methods combine features from different layers of a deep convolutional network, they use the same features extraction for all candidates. Different from the existing methods, we focus on combining multi-layer features in a scale-aware manner to adaptively choose the most suitable feature representation for pedestrians of a certain scale.

\section{Proposed Method}
The overview of the proposed scale-aware multi-resolution (SAM) pedestrian detection framework is illustrated in Figure \ref{fig:long-framework}. Given an input image, RPN generates a pool of pedestrian candidates with an estimated size and a confidence score that will be used as priors for the second stage classification. Pedestrian candidates are grouped according to their sizes, and each group performs feature extraction differently using our proposed SAM method. For small-scaled pedestrian candidates, CNN feature maps at shallow layers, such as $Conv3$ and $Conv4$, are used since CNN feature maps at even deeper layers are considered to be of too low resolution to provide useful information. For large-scaled pedestrian candidates, resolution of feature maps from deeper layers, e.g. $Conv4$ and $Conv5$, are of sufficient resolution and thus, can be utilised. Since pedestrians of different sizes adopt features from different layers of a CNN, their feature representations are of different sizes. PCA is applied to transform the feature representations of different sizes into a fixed length vector. In addition, complementary feature channels are integrated with the CNN features output from SAM and are fed into a boosted forest for classification. A detailed description of each processing block of the proposed framework is given below.

\subsection{Region Proposal Network for Pedestrian Candidates Proposal and Scale Estimation}
The scale of the candidate pedestrians is estimated using the RPN \cite{fasterRCNN2015}, which is a small network built on top of the last convolutional layer of the VGG-16 network. For general object detection \cite{fasterRCNN2015}, a three-scale three-ratio anchor is used to generate 9 proposals at each sliding position. For pedestrian candidates proposal, we use anchors of a single ratio of $\gamma=0.41$ with $9$ scales and fine-tune the RPN \cite{IsfasterRCNNPed} on the Caltech pedestrian benchmark \cite{PedBenchmark-2009CVPR-Dollar}.

Each candidate window output from RPN is associated with information regarding window position, confidence score, and window size. The confidence score is passed to the boosted forest (BF) classifier as the initial detection score. The RPN can generate high quality pedestrian proposals. With $100$ proposals per image, the RPN can achieve $>99\%$ recall at an intersection over union (IoU) of $0.5$, and $>95\%$ recall at an IoU of $0.7$. The candidate windows are sorted by their confidence scores in descending order. At test time, the top-ranked $100$ candidates are passed to the BF classifier for classification, while for training the top-ranked $1,000$ proposals are used. 

\subsection{Scale-Aware Multi-Resolution Features }

Features extracted from different CNN layers represent different levels of abstraction and can be all helpful for pedestrian detection. Features from a deeper CNN layer have stronger representation ability but lower resolution, whereas features from a shallower layer is of higher resolution but weaker representation ability. Using features from unsuitable CNN layers will bring difficulties to the classification process. For instance, when an image of $480\times360$ containing a small pedestrian of $50$ pixels in height is forward to the VGG-16, it will pass through 4 pooling layers and the feature maps at Conv5 will be down-sampled by a factor of $2^{4}$. This leads to the size of active feature maps for the pedestrian being about only $3$ pixels in height. With such low-resolution feature maps, classifiers can hardly discriminate between pedestrians and other irrelevant objects in the scene.

\noindent \textbf{CNN Features Extraction.} CNN features are extracted from the last layer of each convolutional block in the VGG-16 network, i.e., $Conv1\_2$, $Conv2\_2$, ..., $Conv5\_3$. For simplicity, we refer to them as $Conv1$, $Conv2$, ..., $Conv5$. In addition, we also exploit the ``$\grave{a}$ trous'' version of CNN features. The ``$\grave{a}$ trous'' convolution technique is proposed in \cite{SemanticDeepCFRs14} which doubles the feature resolution extracted from $Conv4$ to achieve better semantic segmentation performance. An $\grave{a}$ trous feature map is obtained by dilating the original filter size by a factor of 2 so that the stride of the original feature map can be reduced by $2$. Using the $\grave{a}$ trous convolution enables a higher feature resolution while preserving the same feature representation ability. This is crucial for small object detection. Hence, we also perform experiments on the dilated version of $Conv4$ and $Conv5$ features and refer to them as $Conv4\grave{a}$ and $Conv5\grave{a}$ henceforth. RoI max pooling \cite{fasterRCNN2015} is adopted in order to obtain fix-length CNN feature vectors for candidate region with varying sizes. Unlike in \cite{fasterRCNN2015} where only $Conv5$ features are extracted, we combine multi-resolution feature maps from multiple layers of a CNN.
In order to exploit suitable feature representation for pedestrians of different sizes, we have conducted extensive experiments on two subsets of pedestrians, i.e. small-size and large-size pedestrian sub-dataset, containing pedestrians of height in pixels belonging to the range of $[50,80)$ and $[80,\infty)$, respectively. A comprehensive analysis of using different multi-resolution CNN features is given in Section 4.2.

\vspace*{0.5em}
\noindent \textbf{Multi-resolution Feature Combination using PCA.} Taking the advantage of multi-layer feature aggregation, we utilize multi-resolution CNN features wherein the feature combinations are determined according to the size of the candidate pedestrian. Since CNN features from different layers have different dimension, feature representation for candidates of different size could be of non-uniform length. The PCA algorithm is applied to project these features into fixed length vectors. For a feature combination, we collect a large number of training features from both pedestrian regions and background regions. The eigenvectors corresponding to the top $d$ eigenvalues have been reserved to form a projection matrix for dimension reduction. In this way, the feature representation of pedestrians with varying sizes can be transformed into the same size while preserving their representation ability.

\subsection{SAM+: Enhance SAM using Additional Feature Channels}

The enhanced SAM detector utilizes additional feature channels, i.e. semantic feature channel and edge feature channel to provide complementary information for pedestrian detection. 

We propose an \textbf{RoI histogram pooling} method to extract feature vectors from the additional feature maps for candidate regions of arbitrary size. RoI histogram pooling works by dividing the $H\times W$ RoI window into an $m\times n$ grid of non-overlapping cells so that each cell is of approximate size $(H/m)\times(W/n)$. Differently from RoI max pooling \cite{fasterRCNN2015} which pools the maximum value in each cell into the corresponding output grid cell, we pool out a normalized histogram of values $c\in C$ computed across the feature maps of this cell, i.e. $h_{c}=n_{c}/(m\times n)$ where $n_{c}$ is the number of pixels in this cell belonging to class $c$. Then histogram vectors $h_{c}$ of all cells in this RoI window are concatenated to generate the feature representation of this candidate region. For both RoI max pooling and RoI histogram pooling layer, we use $m=12$ and $n=5$ which is suitable for the pedestrian aspect ratio. We demonstrate in Section 4.2.2 that the features extracted using the proposed RoI histogram pooling perform better than those using RoI max pooling.

\noindent \textbf{Semantic feature channel.} We adopt the recent semantic segmentation method RefineNet \cite{refineNet} to perform pixel-wise semantic labeling and output valuable semantic information for pedestrian detection. RefineNet is trained on the Cityscapes dataset \cite{Cityscapes} to semantically label an image into $C=20$ common classes (e.g. building, tree, sky, road, pedestrian, etc.). We encode the semantic segmentations with the object category index $c$ ($c=1,...,20$) for each pixel as the semantic feature channel. 

\noindent \textbf{Edge feature channel.} Our edge feature channel is encoded using the intensity values of the edge response from the HedNet \cite{HED_edge}. Different from traditional edge detectors such as the Canny, the HedNet can generate semantically meaningful edge maps at object contour.

\subsection{Boosted Forest for Integrated Multi-Resolution Multi-Channel Features }
Boosted Forest (BF) has been widely used in computer vision tasks such as object recognition \cite{HoughForest-2013,DiscrimHoughForest-2013BMVC}, and super-resolution \cite{Tianrui-TIP2015,Tianrui-CVPR2017srhrf+,Junjie-TCSVT} as it can achieve fast and accurate classification.
The flexibility of BF facilitates the combination of multi-layer CNN features without the need for feature amplitude normalization, and is also convenient for the integration of additional feature channels in the SAM+ detector. The confidence scores of the candidate pedestrians output from RPN are passed to BF as preliminary scores for classification.

We adopt the RealBoost algorithm \cite{RealBoost1999} to perform bootstrapping for multi-stage hard negative samples mining.  Our SAM detector performs $6$ stages of bootstrapping passes in addition to the original training phase. The number of weak classifiers used in BF are $64,128,256,512,1024,2048$ for stage $0,1,...,5$ respectively. Initially, the training set consists of all positive examples and $30,000$ negative samples which are randomly sampled from background image regions. For each bootstrapping stage, $5,000$ hard negative samples are collected using the detector obtained from the previous stage. 

We also built a smaller version of SAM, termed \textbf{SAM-Basic}, which allows us to exploit many more different settings in a short training time.\textbf{ }SAM-Basic has $5$ training stages. Each stage has $\{32,64,128,256,512\}$ weak classifiers, respectively. At the first stage, $10,000$ negative samples are randomly sampled and, the number of hard negative samples to be added at each bootstrapping pass is limited to $1,000$.

\section{Experiments and Analysis}
\subsection{Datasets}

\textbf{Caltech.} Caltech-USA dataset \cite{PedBenchmark-2009CVPR-Dollar} and the improved annotations \cite{HowFar-2016} are used for training and evaluation. As in \cite{CompACTDeep,IsfasterRCNNPed,2015scale-awareRCNN}, we use the Caltech10$\times$ training set which is obtained by extracting every $3^{rd}$ frame from the Caltech videos. The testing set contains $4024$ images of size $480\times640$. Following the evaluation of Caltech benchmark \cite{PedBenchmark-2009CVPR-Dollar}, only bounding boxes restricted in the range of $x\in[5, 635]$, $y\in [5, 475]$ are evaluated. The "reasonable" evaluation setting is used which counts for pedestrians of height larger than $50$ pixels and with less than 35\% occlusion.   Evaluations are measured using the log average miss rate (MR) of false positive per image (FFPI) ranging from $10^{-2}$ to $10^{0}$ (\textit{MR}$_{-2}$). For SAM-Basic, the Caltech1$\times$ training set with 4250 images is used and the evaluation is performed on a subset of the Caltech testing set which contains $905$ images.

\vspace*{0.6em}

\noindent \textbf{KITTI.} The KITTI dataset \cite{KITTI-data} is composed of 7,481 images for training and 7,581 images for testing. Since the ground-truth annotations of testing set is not publicly available, we use the training/validation set, as in \cite{whatcanhelpPed,ms-cnn16}, for performance analysis. 
As in KITTI standard, we evaluate our detection methods at three levels of difficulty, i.e. \textquotedblleft Easy\textquotedblright, \textquotedblleft Moderate\textquotedblright and \textquotedblleft Hard\textquotedblright{}, where the difficulty is measure by the minimal height, the occlusion and the truncation of an object. The mean Average Precision (mAP) with  0.5 overlap ratio is used to measure the pedestrian detection performance.

Our implementation is based on the publicly available code for Faster-RCNN \cite{PDollarToolbox,IsfasterRCNNPed} with Caffe \cite{jia2014caffe}. All experiments were performed on a machine with a single GPU TITAN X and a CPU Intel Core i7 $3.4$ GHz.

\subsection{Results and Discussions}
\subsubsection{Scale-Aware Multi-Resolution Convolutional Features}

\begin{figure*}[t]
\begin{centering}
\center\includegraphics[width=1\columnwidth]{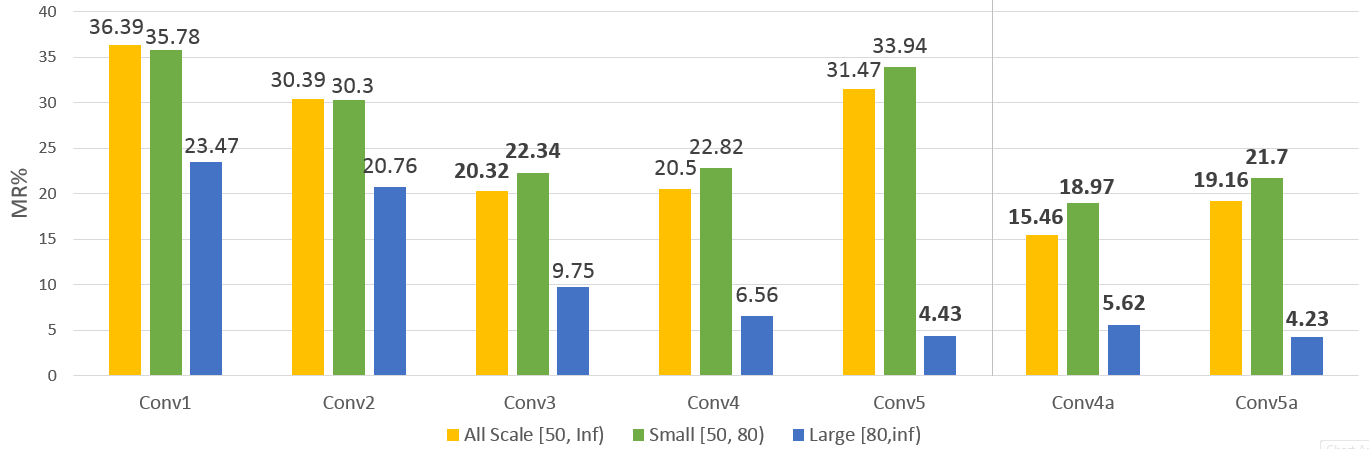}
\par\end{centering}
\caption{Comparison of SAM-Basic results ($MR_{-4}$\%) for "All Scale", "Small" and "Large" pedestrians using feature representation from different CNN layers, the lower the better.}
\label{fig:single-CNNfeat-layer} 
\end{figure*}

We analyse the performance of the SAM-Basic detector using different layers of CNN features and different feature combinations. The experiments were performed under the same parameter settings, except that the CNN features were extracted from different layers.

\noindent \textbf{Single Layer Convolutional Feature for Pedestrians of Difference Scales.} First, we demonstrate that the most suitable features for pedestrians of different scales are from different convolutional layers. We have trained and evaluated SAM-Basic detectors using the "Small" , "Large" and All-scale" pedestrians and compared the results in Figure \ref{fig:single-CNNfeat-layer}. At each time, features from a single convolutional layer, i.e. $Conv1, ...,Conv3, Conv4/ Conv4\grave{a}$ and $Conv5/ Conv5\grave{a}$, are used for training. The log miss rates (MR) of SAM-Basic detectors are averaged over the FPPI range $[10^{-4},10^{0}]$ (\textit{MR}$_{-4}$) since \textit{MR}$_{-2}$ for large-size pedestrians are nearly zero and is difficult to be compared.

From Figure \ref{fig:single-CNNfeat-layer}, we can see that the SAM-Basic detectors perform differently when using features from different CNN layers. More importantly, the best feature representation for small pedestrians is different from that for large pedestrians. Excluding the $\grave{a}$ trous convolutional layers, the best performance for small-size pedestrians is achieved by using $Conv3$ features, while the best of large-size pedestrians is obtained using $Conv5$ features. This verifies that the optimized CNN features for pedestrians of different sizes are from different convolutional layers.

For large-size pedestrians, the lowest miss rates are achieved by using the $Conv5$ and $Conv5\grave{a}$ features. The result indicates that using features from deeper layers which have stronger representation ability is indeed beneficial, as long as the feature resolution is proper for the object to be detected. This inference can be further confirmed by comparing the results between $Conv4$ (or $Conv4\grave{a}$) and $Conv5$ (or $Conv5\grave{a}$). For instance, features from $Conv4$ and $Conv4\grave{a}$ are of the same representation ability, but the results using $Conv4\grave{a}$ features are better because $Conv4\grave{a}$ has doubled feature resolution. 
For small-size pedestrians, results using $Conv1$ and $Conv2$ features are poor because these shallow layers have weak feature representation capability. Performance using $Conv4$ and $Conv5$ features is not satisfied which can be explained by the in-sufficient feature resolution for detecting small pedestrians. Although features from deeper layers have stronger representation ability, these features have too low resolution to let the classifier make good classification. $Conv3$ yields the best result because it is the layer that can best balance feature abstraction level and feature resolution.

\vspace*{0.3em}

\noindent \textbf{Scale-Aware Multi-Resolution Convolutional Features.} According to the results on single layer convolutional features for pedestrians of difference scales, we conduct experiments using combinations of multi-layer CNN features. The features that lead to the best three results (bold in Figure \ref{fig:single-CNNfeat-layer}) are used for combination. The performance of different combinations are shown in Table \ref{tab:feat-layers-Combine}, where the multi-resolution features are shortened for concise (e.g., the combination of features from $Conv3$ and $Conv4\grave{a}$ is denoted as $Conv34\grave{a}$). 

From Table \ref{tab:feat-layers-Combine}, we can see that for large-size pedestrians, the feature combination of $Conv34\grave{a}5\grave{a}$ achieves the lowest miss rate; while for small-size pedestrians the best performance is yielded by $Conv34\grave{a}$. Moreover, by comparing the results between $Conv4\grave{a}5\grave{a}$ and $Conv34\grave{a}5\grave{a}$ for both small and large-size pedestrians, we have an interesting observation: while the incorporation of $Conv3$ features facilitates the detection of small pedestrians, it is on the contrary harmful to the detection of large pedestrians. Therefore, it is not always better to combine more layers of CNN features. As $Conv3$ features are not of the most suitable resolution for large-size pedestrians, combining such features will even degrade the performance.

\begin{table*}[t]
\begin{centering}
\begin{minipage}[b]{0.43\paperwidth}%
\normalsize{
\begin{tabular}{c||c|c|c}
\hline Small & $Conv34\grave{a}$ & $Conv4\grave{a}5\grave{a}$ & $Conv34\grave{a}5\grave{a}$\tabularnewline
\cline{2-4}$[50,80)$ & \textbf{17.55} & 19.07 & 18.22\tabularnewline
\hline
\end{tabular}}%
\end{minipage}
\vfill{}
\vspace{0.5em}
\begin{minipage}[b]{0.43\paperwidth}%
\normalsize{
\begin{tabular}{c||c|c|c}
\hline Large & $Conv4\grave{a}$5 & $Conv4\grave{a}5\grave{a}$ & $Conv34\grave{a}5\grave{a}$\tabularnewline
\cline{2-4}$[80,\infty]$ & 7.49 & \textbf{5.88} & 6.69\tabularnewline
\hline
\end{tabular}}%
\end{minipage}
\par\end{centering}
\vspace*{0.3em}
\caption{Comparison of SAM-Basic result (\textit{MR}$_{-4}$\%) for  small-size (upper) and large-size (lower) pedestrian detection using different combinations of multi-resolution CNN features. \label{tab:feat-layers-Combine}}
\end{table*}

\subsubsection{Comparison between SAM and SAM+ }

Given the results in Table \ref{tab:feat-layers-Combine}, we evaluate our SAM-Basic detector using the best multi-resolution feature combinations for each scale range. That is, for small-size candidates, features are extracted from $Conv3$ and $Conv4\grave{a}$, and for large-size pedestrians, features are extracted from $Conv4\grave{a}$ and $Conv5\grave{a}$. The dimension of $Conv34\grave{a}$ feature is $512+256=786$, whereas the dimension of $Conv4\grave{a}5\grave{a}$ feature is $512+512=1024$. We apply PCA on $Conv4\grave{a}5\grave{a}$ features to reduce their dimension to $786$, so that candidates of different sizes can have uniform feature length. To analyse the influence of feature dimension reduction using PCA, we evaluate the performance of SAM-Basic using the Conv45$\grave{a}$ features before and after dimension reduction. From Table \ref{tab:Comparison-of-PCA}, we can see that there is no performance deterioration in terms of \textit{MR} when the dimension of $Conv45\grave{a}$ features is reduced from $1024$ to $786$.
\vspace*{-0.1em}
\begin{table}[t]
\center%
\begin{tabular}{c|c|c|c}
\hline 
 & Feature dim. & Energy\% & \textit{MR}\%\tabularnewline
\hline 
Original & 1024 & 100 & 17.24\tabularnewline
\hline 
Reduced & 786 & 98.4 & 17.15\tabularnewline
\hline 
\end{tabular}
\vspace*{0.4em}
\caption{Comparison of SAM-Basic results (\textit{MR}$_{-4}$\%) using $Conv45\grave{a}$ features before and after PCA feature reduction.\label{tab:Comparison-of-PCA}}
\end{table}


\begin{table}[t]
\center
\begin{tabular}{c|c|c|c|c}
\hline
{\backslashbox{}{}} 
& All scale  {[}50, $\infty${]}  & Large  {[}80, $\infty${]} & Small  {[}50, 80{)} &Pooling \tabularnewline
\hline 
SAM-Basic & 16.16 & 7.00 & 18.79 &- \tabularnewline
\hline
\multirow{2}{*}{+Semantic} & 15.77 & 7.15 & 17.98 & max.\tabularnewline
\cline{2-5}
 & \textbf{15.14}\cellcolor{lightgray} & \textbf{6.63}\cellcolor{lightgray} & \textbf{16.55}\cellcolor{lightgray} & hist.\cellcolor{lightgray}\tabularnewline
\hline
\multirow{2}{*}{+  Edge} & 15.51 & 7.01 & 17.96 & max.\tabularnewline
\cline{2-5}
 & \textbf{14.73}\cellcolor{lightgray} & \textbf{6.54}\cellcolor{lightgray} & \textbf{15.80}\cellcolor{lightgray} & hist.\cellcolor{lightgray}\tabularnewline
\hline
\end{tabular}
\vspace*{0.4em}
\caption{Performance of SAM+ using semantic and edge feature channels (\textit{MR}$_{-4}$\%, the lower the better), \textquotedblleft max.\textquotedblright{} and \textquotedblleft hist.\textquotedblright{}
indicates RoI max pooling and RoI histogram pooling, respectively.
\label{tab:SAM+}}
\end{table}

The performance of the SAM+ detector using complementary feature channels is given in Table \ref{tab:SAM+}, where \textquotedblleft max.\textquotedblright{} and \textquotedblleft hist.\textquotedblright{} in the last column indicates that the complementary features are extracted using the RoI max pooling and our RoI histogram pooling method, respectively. By comparing the miss rate between \textquotedblleft \textit{max}.\textquotedblright{} and \textquotedblleft \textit{hist}.\textquotedblright , we can see that the proposed RoI histogram pooling performs better than RoI max pooling for semantic and edge feature extraction. We witness there is an overall improvement of $1.02${\%} from the integration of the semantic feature channels. When we look at the fine-grained improvements for different scale ranges, we find that the improvement for small-size pedestrians (i.e., $2.24${\%}) is larger than that for large-size pedestrians (i.e., $0.37${\%}). This indicates that the semantic channel are more helpful for small pedestrians which are usually the hard cases in pedestrian detection. On the basis of using semantic feature channel, the integration of edge feature channel further improves the performance by $0.38${\%}. Again, the detection rate of on small-size pedestrians benefits more than that of the large-size pedestrians.


\vspace{-1em}
\subsubsection{Comparison with State-of-the-art Pedestrian Detection Methods}

\begin{figure}[t]
\centering{}%
\hfill{}
\begin{minipage}[b]{0.47\columnwidth}%
{\includegraphics[width=1\columnwidth]{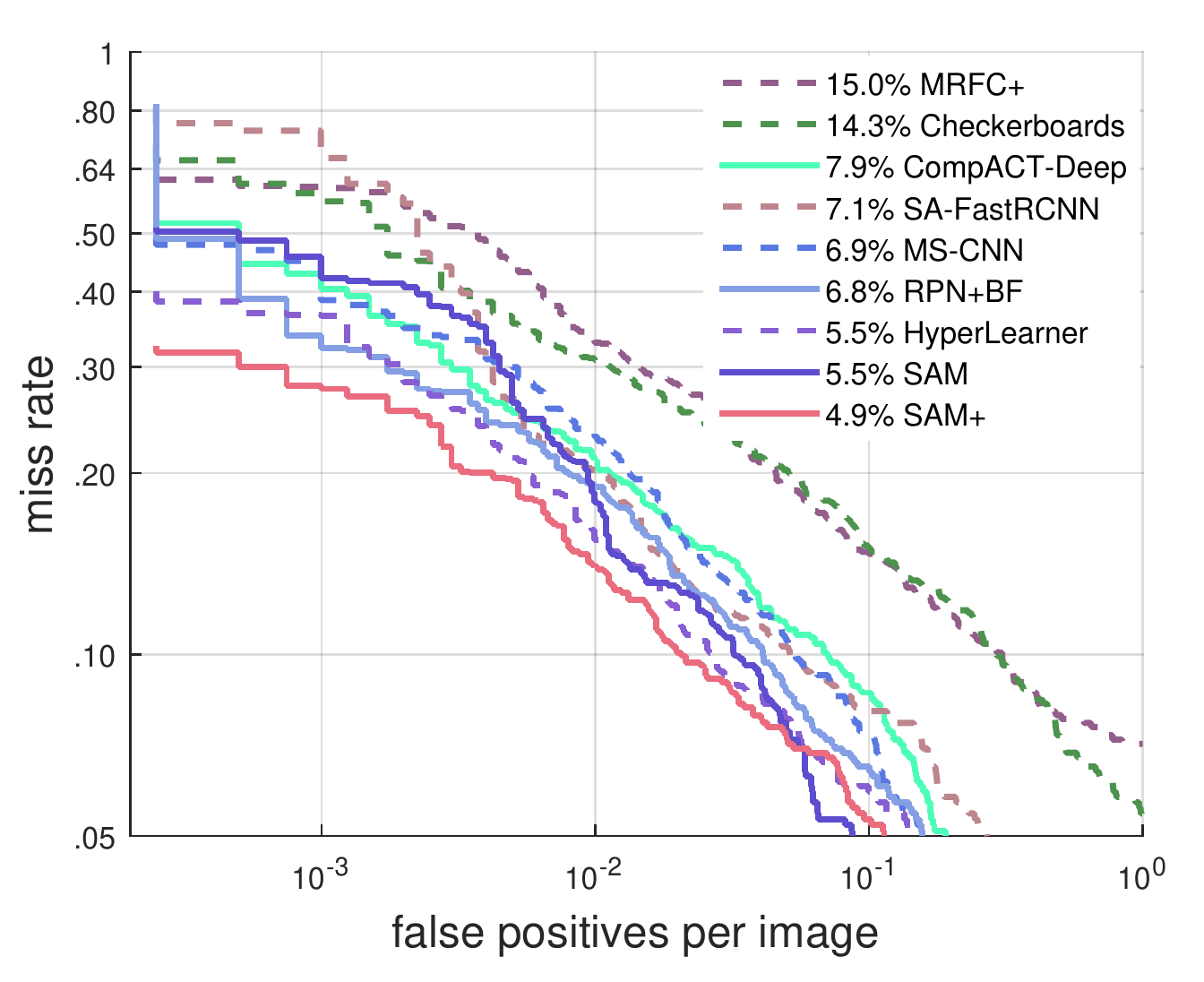}}%
\end{minipage}%
\hfill{}
\begin{minipage}[b]{0.47\columnwidth}%
{\includegraphics[width=1\columnwidth]{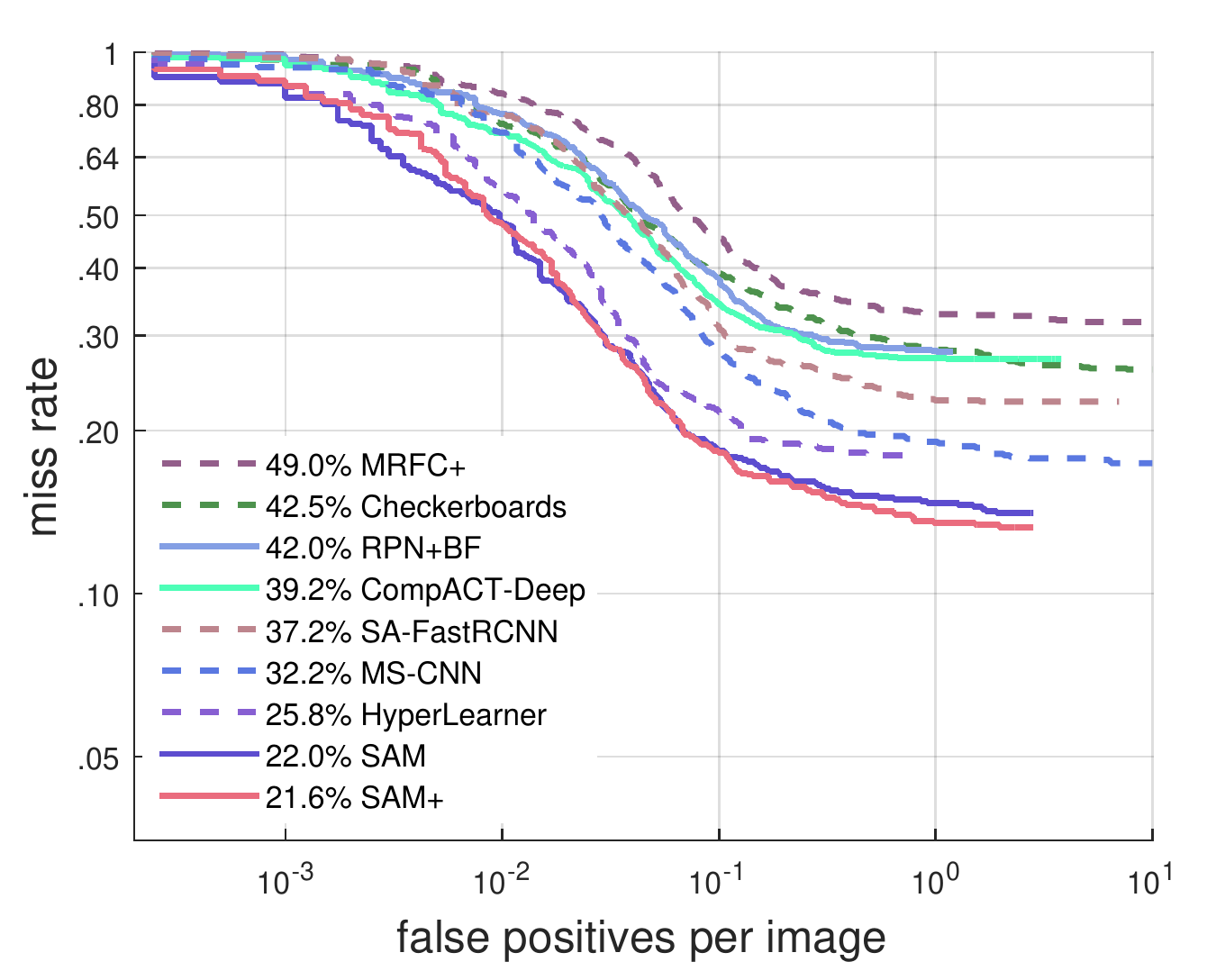}}%
\end{minipage}%
\hfill{}
\caption{Comparison of results (\textit{MR}$_{-2}$) on the Caltech test set evaluated using IoU 0.5 (left) and 0.7 (right), respectively.}
\label{fig:Comparison-of-StateofArts}
\end{figure}

\textbf{Caltech.}  In Figure \ref{fig:Comparison-of-StateofArts}, our SAM and SAM+ pedestrian detectors are compared with the state-of-the-art pedestrian detection methods, namely,  Checkerboards \cite{FilterChannal-2015}, MRFC \cite{SemanticChannel}, CompACTDeep \cite{CompACTDeep}, SA-FastRCNN \cite{2015scale-awareRCNN}, MS-CNN \cite{ms-cnn16} , RPN+BF \cite{IsfasterRCNNPed},  and HyperLearner \cite{whatcanhelpPed}. Under the evaluation setting of IoU is $0.5$ (Figure \ref{fig:Comparison-of-StateofArts} (left)), the performance of SAM is on par with that of the latest HyperLearner method \cite{whatcanhelpPed} even without using additional features. Our SAM+ detector has achieved a MR of $4.9$\% which outperforms the current state-of-the-art by $0.6$\%. Under a stricter evaluation condition of IoU is $0.7$ (Figure \ref{fig:Comparison-of-StateofArts} (right)), our proposed SAM and SAM+ method outperform all existing pedestrian detection methods with a larger margin and the performance of SAM+ is $0.4$\% better than that of SAM. This indicates that our proposed method not only achieves lower miss rate, but also obtains detection with more precise position.
\vspace{1em}

\noindent \textbf{KITTI.} Table \ref{KITTI_Table} shows the pedestrian detection results evaluated on the KITTI dataset. Again, our SAM method has competitive accuracy compared to the latest pedestrian detector \cite{whatcanhelpPed} without using additional feature channels. Under the "hard" evaluation setting where pedestrians of $25$ pixels in height and heavily occluded are counted, SAM outperforms HyperLearner \cite{whatcanhelpPed} by 4.67\%. The SAM+ method further improves the performance of SAM under all three evaluation settings.

\begin{table}[t]
\center
\begin{tabular}{c|p{1.4cm}<{\centering}|p{1.4cm}<{\centering}|p{1.4cm}<{\centering}}
\hline
Method & Moderate & Easy & Hard\tabularnewline
\hline
Faster-RCNN \cite{IsfasterRCNNPed}& $71.05$ & $76.00$ & $62.08$\tabularnewline
\hline
MS-CNN \cite{ms-cnn16} & $72.26$ & $76.38$ & $64.08$\tabularnewline
\hline
HyperNet \cite{whatcanhelpPed} & 72.23  & 77.96 & 63.43\tabularnewline
\hline
HyperLearner \cite{whatcanhelpPed} & $72.51$ & $78.51$  & $63.24$\tabularnewline
\hline
SAM [Our] &  \textbf{74.07} &  \textbf{78.80}  &  \textbf{67.91}\tabularnewline
\hline
SAM+ [Our] & \textbf{74.25} & \textbf{78.92} & \textbf{68.40}\tabularnewline
\hline
\end{tabular}
\vspace*{0.4em}
\caption{Comparisons of pedestrian detection results (mAP\%) on KITTI.
\label{KITTI_Table}}
\end{table}

\subsection{Conclusions}

In this paper, we proposed a scale-aware multi-resolution (SAM) pedestrian detection framework which exploits different combination of multi-resolution CNN features for pedestrian candidates of different scales. Through extensive experiments, we found that for pedestrians of different scales, the features that can best balance feature abstraction level and resolution are from different convolutional layers. It is not always better to combine more layers of CNN features. Using features of unsuitable resolution will bring difficulties to the detector in the classification stage, and thus is harmful for detection accuracy. We also proposed the enhanced SAM+ detector which makes use of the additional feature channels as complementary information for pedestrian detection. Relying on the additional cues, some ambiguous pedestrian hypotheses that may be difficult to classify using the CNN features can be discriminated with the proposed method. Experiments indicate that the proposed detectors achieve state-of-the-art performance.

\subsection{Acknowledgment}
This work was supported by the EU H2020 TERPSICHORE project ``Transforming Intangible Folkloric Performing Arts into Tangible Choreographic Digital Objects'' under the grant agreement 691218.

\bibliography{BMVC18}
\end{document}